%% file: arxiv.tex
\documentclass{article}

\usepackage[preprint]{neurips}

\usepackage{natbib}
\setcitestyle{numbers,square}

\usepackage[utf8]{inputenc} 
\usepackage[T1]{fontenc}    
\usepackage{hyperref}       
\usepackage{url}            
\usepackage{booktabs}       
\usepackage{amsfonts}       
\usepackage{nicefrac}       
\usepackage{microtype}      
\usepackage{xcolor}         
\usepackage{subcaption}

\usepackage{amsmath}
\usepackage{xspace}
\usepackage{amssymb}
\usepackage{graphicx}
\usepackage{multirow}
\usepackage{enumitem}
\usepackage{ulem}
\usepackage{float}
\usepackage{makecell}
\usepackage{caption}
\usepackage{threeparttable}
\usepackage{fancyhdr}
\newcommand{\eg}{\textit{e.g.,}\xspace}
\newcommand{\etal}{\textit{et al.}\xspace}
\newcommand{\paratitle}[1]{\vspace{0.8ex}\noindent \textbf{#1}}

\newcommand{\modelname}{Tele-FLM\xspace}

\title{\modelname Technical Report}

\author{
  Xiang Li\textsuperscript{1\textdagger}, 
  Yiqun Yao\textsuperscript{1\textdagger},
  Xin Jiang\textsuperscript{1\textdagger}, 
  Xuezhi Fang\textsuperscript{1\textdagger}, 
  Chao Wang\textsuperscript{2\textdagger},
  Xinzhang Liu\textsuperscript{2\textdagger},\\
  \textbf{
  Zihan Wang\textsuperscript{2},
  Yu Zhao\textsuperscript{2}, 
  Xin Wang\textsuperscript{2},
  Yuyao Huang\textsuperscript{2}, 
  Shuangyong Song\textsuperscript{2},
  Yongxiang Li\textsuperscript{2},
  }\\
  \textbf{
  Zheng Zhang\textsuperscript{1}, 
  Bo Zhao\textsuperscript{1}, 
  Aixin Sun\textsuperscript{3}, 
  Yequan Wang\textsuperscript{1$*$},
  Zhongjiang He\textsuperscript{2$*$},
  } \\
  \textbf{
  Zhongyuan Wang\textsuperscript{1},
  Xuelong Li\textsuperscript{2},
  Tiejun Huang\textsuperscript{1}
  }\\ \\
  $^{1}$Beijing Academy of Artificial Intelligence, Beijing, China\\
  $^{2}$Institute of Artificial Intelligence (TeleAI), China Telecom Corp Ltd, China\\
  $^{3}$School of Computer Science and Engineering, Nanyang Technological University, Singapore
}

\begin{document}

\maketitle

\renewcommand{\thefootnote}{\fnsymbol{footnote}}
\footnotetext[2]{Indicates equal contribution.}
\footnotetext[1]{Corresponding authors.}
\renewcommand{\thefootnote}{\arabic{footnote}}

\begin{abstract}
Large language models (LLMs) have showcased profound capabilities in language understanding and generation, facilitating a wide array of applications. However, there is a notable paucity of detailed, open-sourced methodologies on efficiently scaling LLMs beyond 50 billion parameters with minimum trial-and-error cost and computational resources.  In this report, we introduce \modelname (aka FLM-2), a 52B open-sourced multilingual large language model that features a stable, efficient pre-training paradigm and enhanced factual judgment capabilities. \modelname demonstrates superior multilingual language modeling abilities, measured by BPB on textual corpus. Besides, in both English and Chinese foundation model evaluation, it is comparable to strong open-sourced models that involve larger pre-training FLOPs, such as Llama2-70B and DeepSeek-67B. 
In addition to the model weights, we share the core designs, engineering practices, and training details, which we expect to benefit both the academic and industrial communities.

\end{abstract}

\section{Introduction}
\label{sec:intro}

Large Language Models (LLMs) have been considered a remarkable approach for unsupervised learning, utilizing extensive data to achieve significant advancements. Large models based on decoder-only Transformers \cite{vaswani2017attention, radford2019language} have demonstrated strong abilities on language understanding, generation, and in-context learning \cite{gpt3}, \etal. Through downstream supervised fine-tuning (SFT) and task-specific alignments~(\eg Reinforcement Learning from Human Feedback, RLHF) \cite{instructgpt}, LLMs have led to significant progress in the development of dialogue assistant applications with their human-level multi-turn interaction capabilities \cite{GPT-4}. Furthermore, LLMs have demonstrated complex cognitive abilities as reflected by code interpretation and completion \cite{wizardcoder}, mathematical problem-solving \cite{step-by-step}, logical reasoning \cite{cot}, and agent-like actions \cite{rt-2}. Recently, LLMs have also shown potential to facilitate a unified sequence-to-sequence modeling paradigm for multimodal learning by treating image, video, and audio signals all as token sequences \cite{gemini, videopoet}. This positions LLMs as pivotal for progress towards Artificial General Intelligence (AGI) \cite{sparks}.

Inspired by the superior performances of proprietary applications \cite{GPT-4, claude}, a plethora of open-sourced LLMs has been publicly available for both the English \cite{llama, llama-2, falcon, mistral,gemma} and Chinese \cite{baichuan, qwen, deepseek, flm101b} communities. The open-sourced models typically vary in size from 7B to 70B parameters, with their performances improving with model sizes and training FLOPs, which is described as scaling laws \cite{scaling-law, scaling-law-2}. Open LLMs can be classified into foundation language models, SFT models, and RLHF models.

Despite the growing prevalence and impressive evaluation performances, the high computational cost remains the major challenge in LLM development. In this study, we focus on alleviating the excessive computation by establishing a model-producing pipeline that streamlines the hyperparameter searching process, minimizes trial-and-error, and reduces restarts in training. For instance, the Llama technical report \cite{llama} assumed the use of around 2,048 A100 GPUs for 5 months, while a single Llama-65B training trial spanned only 21 days, constituting only 14\% of the total GPU time. It indicates that open-source endeavors of pre-training LLMs may undergo redundant trial-and-error cycles that may consume enormous computational resources. In contrast, in this work, we reduce the total time cost due to restarts and trial-and-error to negligible levels. We believe that sharing our detailed techniques, engineering practices, and training dynamics \cite{olmo}, especially for LLMs exceeding the 50B scale, could benefit the community as well as contribute to green AI.

In this report, we introduce \modelname (aka FLM-2), an open multilingual LLM with 52 billion parameters, which is pre-trained from scratch on a 2.0 trillion token corpus comprising texts from English, Chinese, and various other languages. 
\modelname inherits and extends the low carbon techniques and fact-enhancing pre-training objectives from the FLM family \cite{flm101b}. The training of \modelname has encountered no instability issue except hardware failures through the completed 2T tokens, and remains ongoing for more data. In addition to the model checkpoints, we release the details of data composition, model architecture, hyperparameter searching, and the full pre-training dynamics.

We evaluate \modelname across multiple English and Chinese benchmarks. Regarding English language modeling, \modelname has better Bits-Per-Byte (BPB) than Llama2-70B \cite{llama-2}, demonstrating strong compression capabilities. The model also achieves lower BPB than Llama3-70B \cite{llama3} and Qwen1.5-72B \cite{qwen} on Chinese corpora, showcasing its multilingual nature. With fewer English training tokens and smaller models, \modelname matches Llama-65B and is comparable to Llama2-70B in English foundation model evaluation.
As for Chinese foundation model evaluation, \modelname matches the overall performance of larger multilingual models trained with a similar amount of data (\eg DeepSeek-67B \cite{deepseek}). On certain tasks, it surpasses larger models trained with significantly more data (\eg Qwen1.5-72B).

The remainder of this report is structured as follows: Section \ref{sec:pre-training-data} delves into the specifics of pre-training data processing. Section \ref{sec:details} details our model architecture, tokenizer, infrastructures, training techniques, and hyperparameters.  
In Section \ref{sec:dynamics-and-bpb}, we illustrate the pre-training dynamics and conduct BPB-based evaluation and analysis. 
Benchmark evaluation in both English and Chinese are provided in Section \ref{sec:benchmark-evaluation}. 
Section \ref{sec:discuss} discusses some common issues and lessons learned. 
Section \ref{sec:related} reviews related literature. We conclude our work and look to the future in Section \ref{sec:con}.

\section{Pre-training Data}
\label{sec:pre-training-data}
Our training dataset comprises a variety of domains, as detailed in Table \ref{train_data_stat}. We build a custom pipeline on spark cluster for massive data processing and apply custom functions to each subset. The pipeline includes text extraction from HTML/WARC, cleaning and paragraph-level deduplication with heuristic rules, model-based quality filtering and document-level deduplication with MinHash \cite{minhash} algorithm. We obtain 2T tokens after all the procedures, and the distribution ratio between English and Chinese data is roughly 2:1. We incorporate more English data because of its higher quality, especially regarding the WebText domain. Additionally, in line with the methodology of GPT-4, we collected some instruct data and incorporated it into our pre-training data after removing the test sets of common datasets using the strict n-gram-based method. We deliberately avoid ``training on the test set'' or any other benchmark-oriented trick.
\input{tables/train_data_stat}

\paratitle{WebText.}
CommonCrawl\footnote{\url{https://commoncrawl.org/}.} is often considered to be a repository containing diverse human experience and rich knowledge~(especially long-tail knowledge). However, the high-quality sources in CommonCrawl are primarily concentrated in the English segment, with the Chinese content exhibiting relatively lower information density and quality. We use the latest CommonCrawl dumps from RedPajama \cite{together2023redpajama} and incorporate WudaoCorpora \cite{wudaocorpora} and similar Chinese-specific datasets together to form a large web-text dataset. We apply custom heuristic rules and a FastText \cite{joulin2016fasttext} classifier to filter out low-quality content, cross-deduplicate for each language, and up-sample/down-sample each subset with regard to data quality. The ratio of English to Chinese is approximately 2:1.

\paratitle{Code.}
We incorporate multiple Github-like code datasets and post-process it to filter-out low quality and duplicated content. Simultaneously, we carefully assembled and curated a well-formed markdown dataset comprising Chinese technical articles.

\paratitle{Book.}
We collect books from various sources in both English and Chinese, such as Redpajama \cite{together2023redpajama} and Gutenberg\footnote{\url{https://www.gutenberg.org/}.}, among others. We develop a series of cleaning steps to remove redundant formatting, garbled text, formula errors, duplicated paragraphs, and other unwanted content from the books. After interleaved deduplication on document level, we finally obtain a high-quality book dataset. The ratio of English to Chinese is nearly 1:1.

\paratitle{WorldKnowledge.} 
To enrich the model's knowledge base and common sense, we add Wikipedia dumps\footnote{\url{https://dumps.wikimedia.org/}.} from 2024 period to our training set, covering 22 languages: bg, ca, cs, da, de, en, es, fr, hr, hu, it, ja, nl, pl, pt, ro, ru, sl, sr, sv, uk, zh. We first process these dumps via Online Language Modelling Dataset Pipeline \cite{thrush2022pipeline} to clean up format; then a meticulous multi-lingual cleaning function is applied to remove reference and subsequent content, which tend to be irrelevant to the main text.

\paratitle{QA.}
We use StackExchange dataset provided by RedPajama-Data \cite{together2023redpajama}. Furthermore, similar Chinese datasets are collected and incorporated into the training after filtering out those QA pairs with low information content. The ratio of English to Chinese in this subset is roughly 1:2.

\paratitle{AcademicPaper.}
We use arxiv dataset collected and processed by RedPajama-Data. This dataset is processed following a Llama-like procedure, which mainly focuses on clearing useless or redundant formats for better language modeling.

\paratitle{Profession.}
To enhance the model's capacity in various professional fields, we decide to include some specific domains in our dataset, including medical, law, patent, and math. Some subsets are from open-source data, such as Wanjuan-Patent \cite{wanjuan} and MathGLM \cite{mathglm}. We post-process each subset independently to address formatting issues, private information disclosure, \etal.

\paratitle{ClassicalChinese.}
In order to improve the model's understanding of traditional Chinese culture and its capability in classical Chinese, we carefully collect classic Chinese ancient books and poetry. These materials are more credible than those found in web texts; therefore, we assign them a larger weight during sampling.

\section{Pre-training Details}
\label{sec:details}
\input{tables/model_architecture_parameters}

\subsection{Model Architecture}

We adapt the architecture of FLM-101B \cite{flm101b} as a backbone with several modifications. FLM-101B follows the standard GPT-style decoder-only transformer architecture \cite{radford2019language}, with pre-normalization and adds a LayerNorm to the last layer's output. Meanwhile, we apply scalar multipliers to: (1) the output of the word embedding layer and (2) the final output hidden states before softmax. We leave these multipliers tunable in pre-training to control the numerical flow. For example, the output multiplier may benefit training by modulating the entropy of the vocabulary distribution.

Building on FLM-101B, we further optimize the model structure for \modelname. Specifically, We use RMSNorm \cite{RMSNorm} for normalization and SwiGLU \cite{swiglu} for the activation function. We roll back to use Rotary Positional Embedding~(RoPE) \cite{rope} without Extrapolatable Position Embedding~(xPos) \cite{xpos}, untie the embedding layer with language modeling head, and disable linear bias in the attention and all MLP modules. 
One mini version named $\text{\modelname}_{\mu\text{P}}$ is used to search hyper-parameters here. Table~\ref{tab:model_architecture_parameters} details the architecture of both \modelname and $\text{\modelname}_{\mu\text{P}}$.

\input{tables/tokenizer_compression_rate}

\subsection{Tokenizer}

The key to training a text tokenizer is to make a better trade-off between compression ratio and vocabulary size. English-focused tokenizers like GPT-4 or previous Llama series often underperform in compressing Chinese text. In order to guarantee \modelname's text compression ratio within Chinese while maintaining performance under multilingual setting, we train a tokenizer that aligns closely with the pre-training data distribution. We sample 12 million diverse text samples from our pre-training dataset as the tokenizer's training dataset, including multilingual texts with a primary focus on Chinese and English, code snippets, classical Chinese literature, and mathematical content. We train the tokenizer with Byte-level BPE (BBPE) algorithm \cite{bbpe}.

Table~\ref{tab:tokenizer_compression_rate} details the tokenizers of \modelname, GPT-4, and the Llama family. The tokenizer of \modelname outperforms GPT-4 and Llama series in both Chinese and Classical Chinese and is comparable with their performances in English, code, and multilingual content. In math, our tokenizer aligns with Llama2 while slightly trailing GPT-4. Overall, \modelname tokenizer showcases a superior compression ratio for Chinese text and satisfactory performance in English. While slightly behind Llama3, \modelname outperforms other approaches on average compression ratio by a large margin.

\subsection{Cluster Hardware}

\modelname is trained on a cluster of 112 A800 SXM4 GPU servers, each with 8 NVLink A800 GPUs and 2TB of RAM. The nodes have heterogeneous CPU architectures: 96 nodes with Intel 8358 (128$\times$ 2.60GHz) CPUs and 16 nodes with AMD 7643 (96$\times$ 2.30GHz) CPUs. All nodes are interconnected via InfiniBand (IB). The training process lasts around two months, including downtime due to unexpected factors. As a comparison of infrastructures, Llama3 \cite{llama3} is pre-trained on at least 49,152 Nvidia H100 GPUs (in contrast to our 896$\times$ A800). Meta also claims to have the equivalent of 600k H100 GPUs for future computing power\footnote{\url{https://www.instagram.com/reel/C2QARHJR1sZ/?hl=en}.}. With this significant gap in total resources, computational efficiency and success rate are critical for average entities.

\subsection{Parallelism}

\modelname utilizes 3D parallel training, combining the prevailing methodologies: data parallelism, tensor parallelism, and pipeline parallelism.

Data parallelism~\cite{parallel-computation} is a well-established distributed training method, in which the samples in a batch are partitioned and distributed across multiple devices and processed simultaneously. No inter-device communication is involved in the forward and backward computation, while the gradient is aggregated at the end of each step.

Tensor parallelism~\cite{DBLP:journals/corr/abs-1909-08053} splits specific neural network tensors across multiple devices and computes via inter-device communication. In \modelname training, tensor parallelism is mainly applied to the attention and feed-forward modules.

Excessive use of tensor parallelism may escalate GPU communication overheads and reduce the training speed. To alleviate this, we integrate pipeline parallelism~\cite{DBLP:journals/corr/abs-2104-04473} that partitions the model at the layer level.

3D parallelism incorporates these parallel approaches, prioritizing allocation of tensor parallelism groups with higher communication overheads to the same node, thereby maximizing intra-node communication and minimizing inter-node communication. The parallel training setup for \modelname is a mixture of 4 tensor parallel, 2 pipeline parallel, and 112 data parallel.

Additionally, we partition inputs to the Transformer's LayerNorm and Dropout layers along the sequence length dimension with sequence parallelism~\cite{korthikanti2022reducing}, yielding further GPU computational and memory savings.
Furthermore, we utilize Distributed Optimizer module from Megetron-LM\footnote{\url{https://github.com/NVIDIA/Megatron-LM}.}~\cite{DBLP:journals/corr/abs-1910-02054} with optimization. This optimizer further reduces GPU memory consumption by partitioning optimizer states with larger memory footprints across the data parallel dimension.

\subsection{Hyperparameter Search}

Effective hyperparameter tuning may accelerate the loss reduction and ensure convergence, making it crucial for model training. However, the high cost of training large models often renders exhaustive grid searches impractical. Hence, we employ $\mu$P \cite{TP5} for optimal parameter search. The Tensor Programs theories \cite{TP4,TP4b} reveal universal relations in the training dynamics across a series of models, with their widths approaching infinity. For certain hyperparameter classes, this leads to a parameterized mapping for their optimal values between small and large widths. Generally, under $\mu$P transfer, wider models will consistently achieve lower loss than narrower ones when trained on identical data \cite{TP5}. Consequently, if a narrow model converges, its wider counterparts will always converge.

Based on this approach, we set a small model, namely $\text{\modelname}_{\mu\text{P}}$, for grid search purpose. As demonstrated in Table~\ref{tab:model_architecture_parameters}, this small model's architecture is different from \modelname only in width. With a fixed layer number of 64 and attention head dimension of 128, we reduce the hidden size to 512. This modification results in 4 attention heads and a feed-forward hidden size of 1344. Due to its smaller size, $\text{\modelname}_{\mu\text{P}}$ allows for significantly more experimental runs within fixed time and resource constraints.

We search 7 hyperparameters: Learning Rate for vector-like and matrix-like weights, the Minimum Learning Rate at the end of the schedule, the initialization Standard Deviation for vector-like and matrix-like weights, the scaling factor for the embedding layer~(namely Input Mult), and the scaling factor for the output hidden state in the final layer~(namely Output Mult). For the definitions of vector/matrix-like weights and the $\mu$P transferring formula we apply, please refer to \cite{mu-scaling} and \cite{TP5}. We use truncated normal distribution for model initialization.

\begin{figure}[h]
    \begin{subfigure}[b]{0.42\textwidth}
        \centering
        \includegraphics[width=1.2\textwidth]{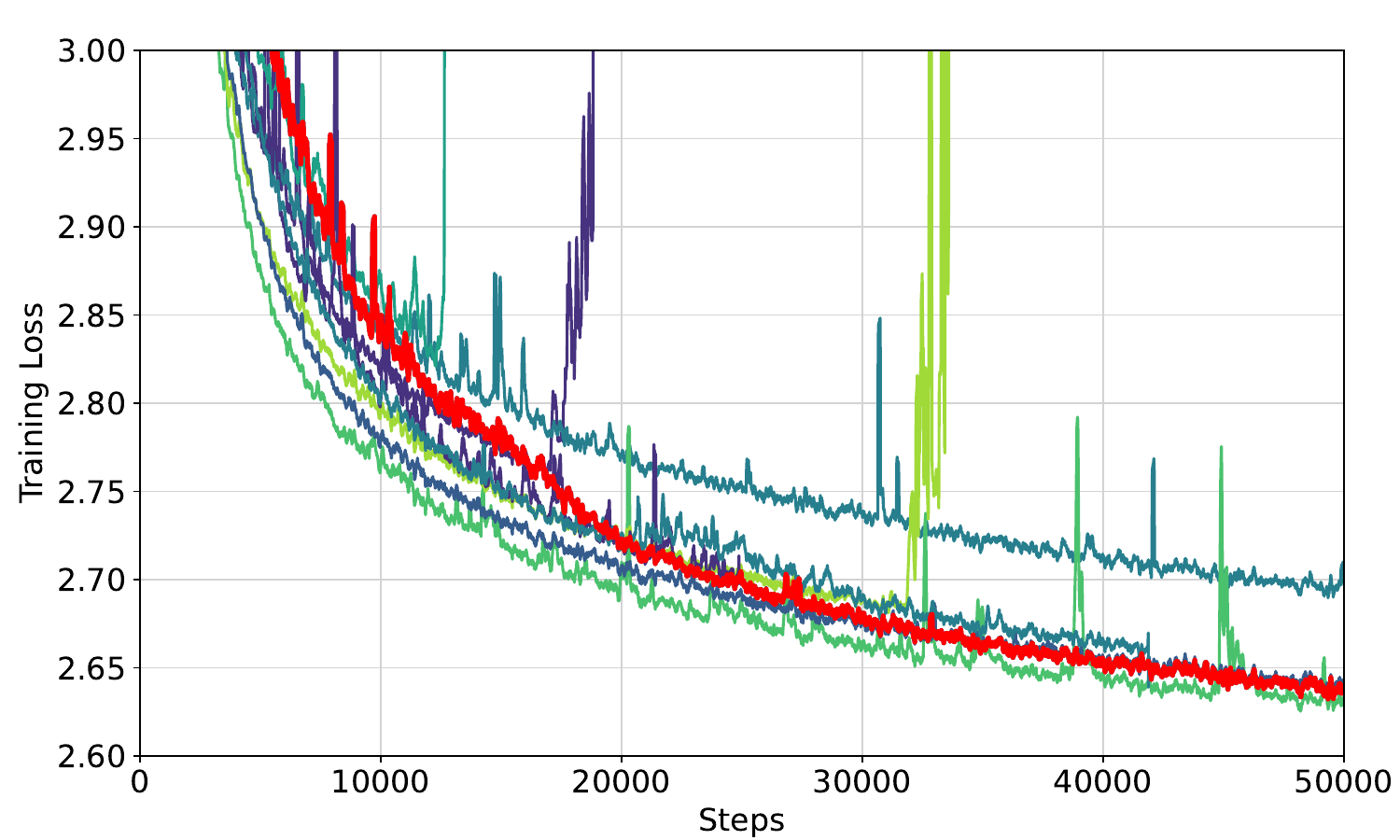}
        \caption{Loss curves for grid search.}
        \label{fig:mup_loss}
    \end{subfigure}
    \hspace{1cm} 
    \begin{subfigure}[b]{0.42\textwidth}
        \centering
        \includegraphics[width=1.18\textwidth]{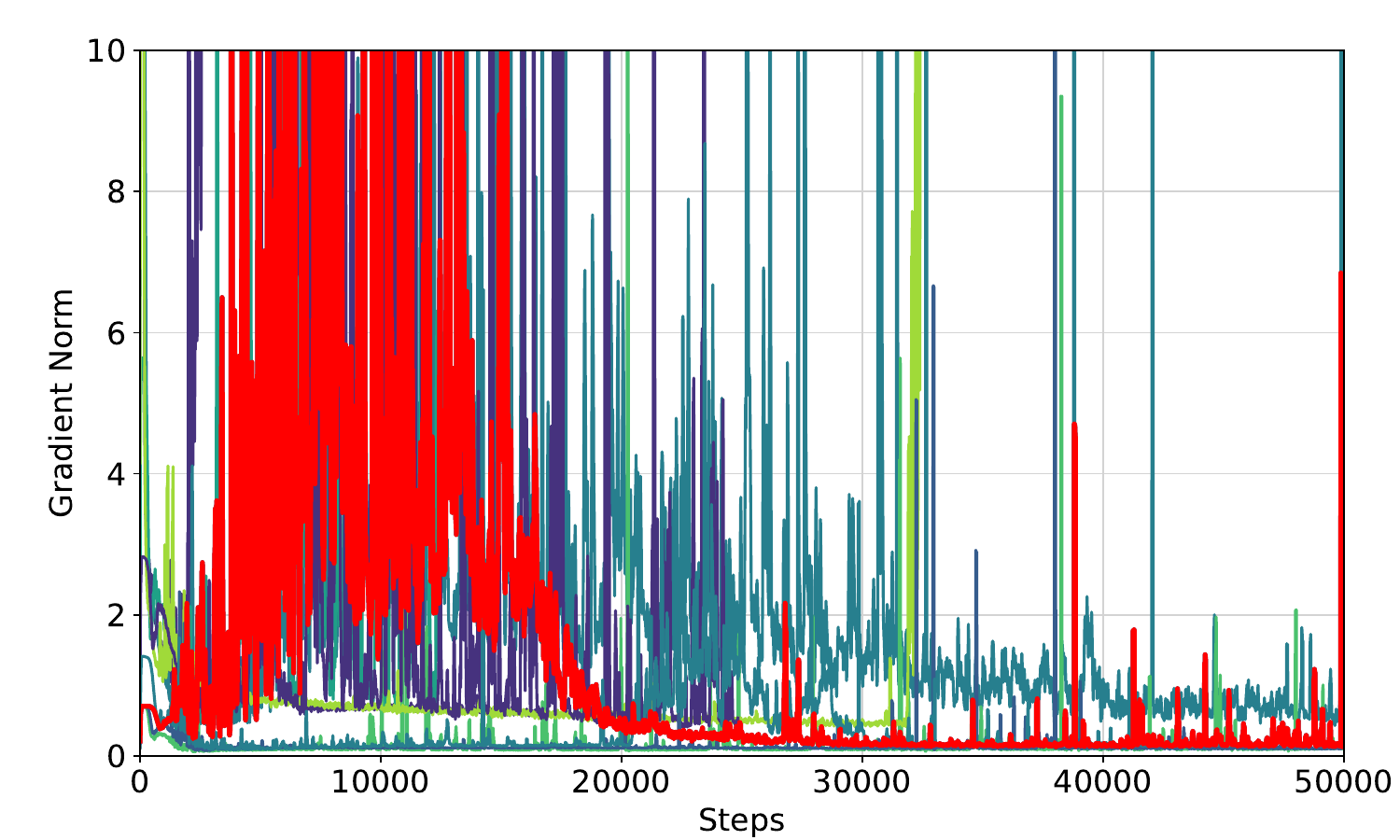}
        \caption{Gradient norm curves for grid search.}
        \label{fig:mup_grad}
    \end{subfigure}
    \caption{Experimental curves of hyperparameter search based on $\mu$P.}
    \label{fig:mup_exps}
\end{figure}

Figure~\ref{fig:mup_exps} illustrates the loss and gradient norm dynamics of 9 hyperparameter combinations for the grid search, which are selected based on our prior knowledge of model configurations. We choose the hyperparameters represented by the red line for final training after assessing the rate of loss decrease, trend stability, and gradient norm stability. Using $\mu$P, we derive the optimal hyperparameter configuration for the final 52B model based on this searched result, which is detailed in Table~\ref{tab:hyperparameters}. A more fine-grained search can be conducted with expanded time and budgets.

\input{tables/hyperparameters}

\section{Loss Dynamics and BPB Evaluation}
\label{sec:dynamics-and-bpb}

\begin{figure}[h]
    \begin{subfigure}[b]{0.3\textwidth}
        \centering
        \includegraphics[width=1\textwidth]{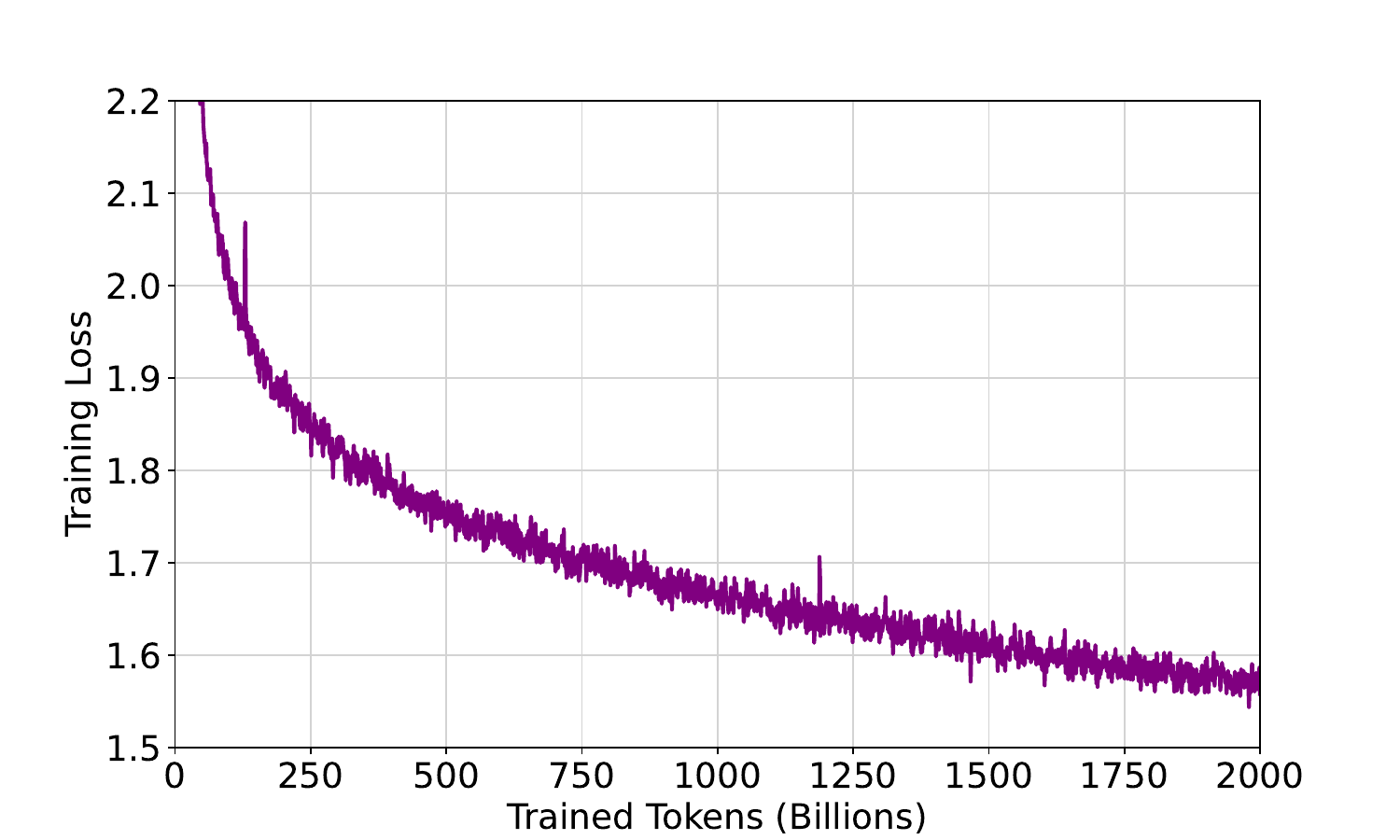}
        \caption{Training loss curve.}
        \label{fig:main_loss_train}
    \end{subfigure}
    \hspace{0.2cm} 
    \begin{subfigure}[b]{0.3\textwidth}
        \centering
        \includegraphics[width=1\textwidth]{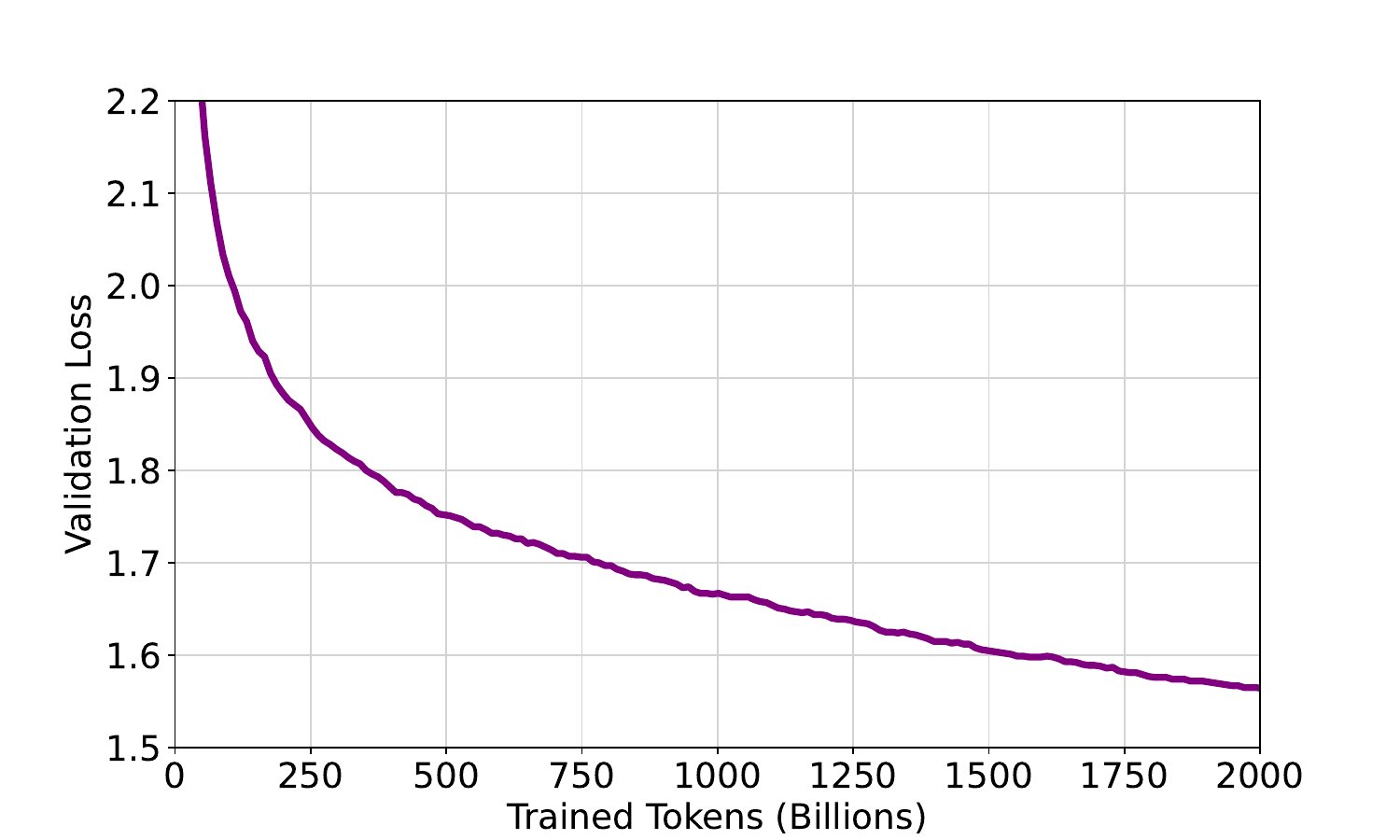}
        \caption{Validation loss curve.}
        \label{fig:main_loss_valid}
    \end{subfigure}
    \hspace{0.2cm} 
    \begin{subfigure}[b]{0.3\textwidth}
        \centering
        \includegraphics[width=1\textwidth]{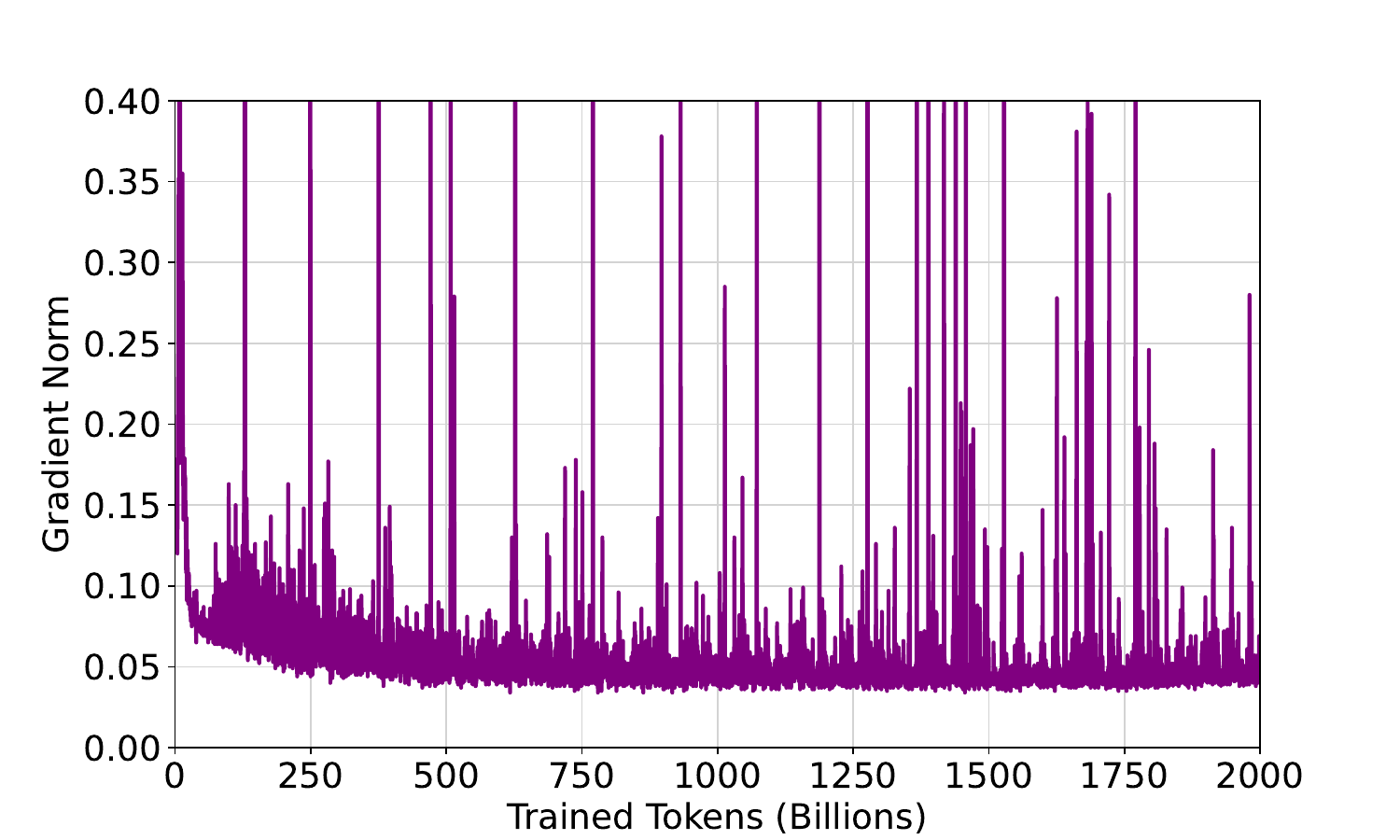}
        \caption{Training gradient norm curve.}
        \label{fig:main_grad_norm}
    \end{subfigure}
    \caption{Pre-training curves for \modelname w.r.t. amount of data in billion tokens.}
    \label{fig:main_loss}
\end{figure}

\begin{figure}[t]
    \centering
    \includegraphics[width=0.99\textwidth]{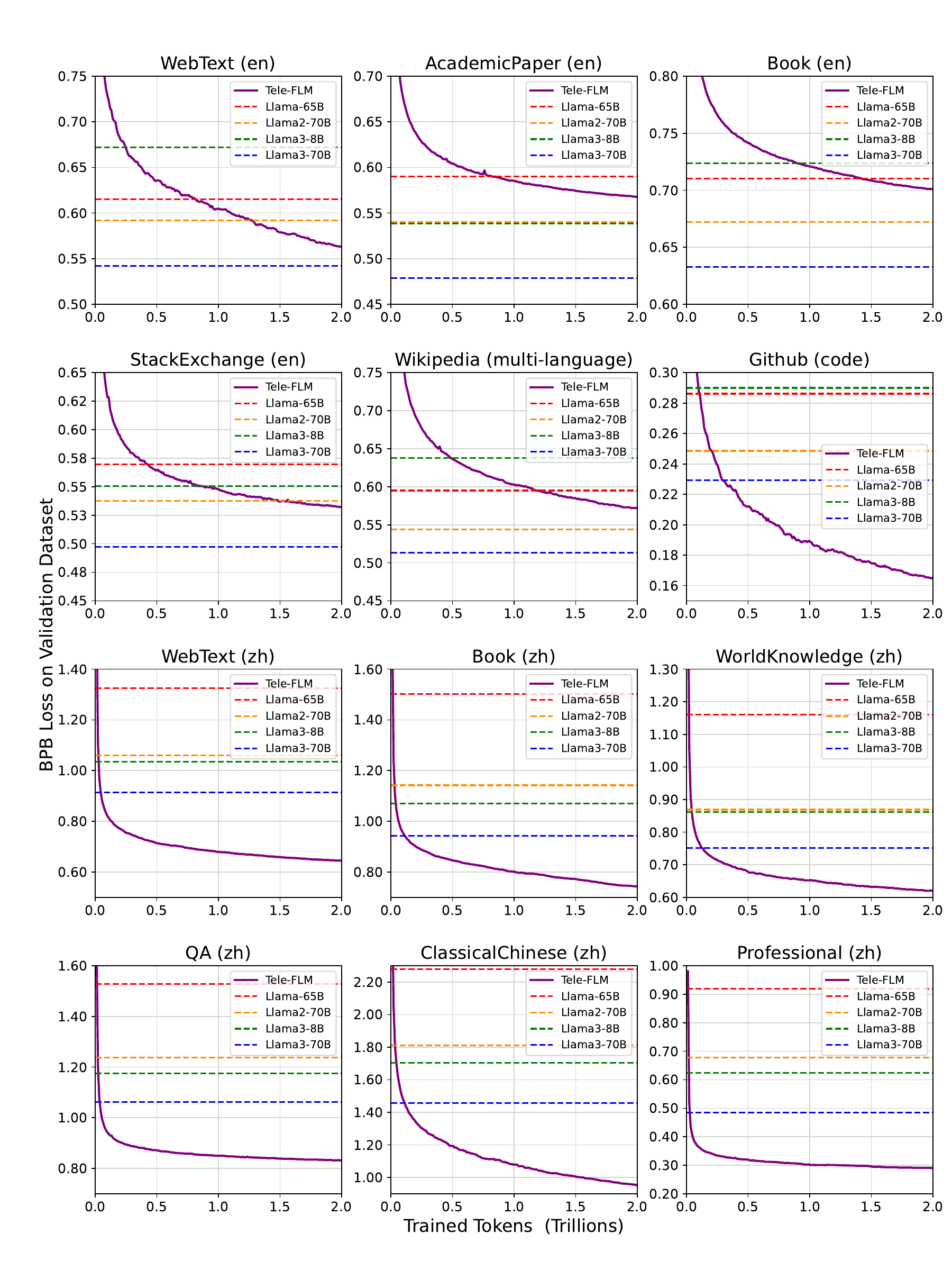}    
    \caption{BPB curves of \modelname on representative English (en), Chinese (zh), multi-language, and code validation datasets, compared with Llama series.}
    \label{fig:valid_bob_loss_with_llama_4x3}
\end{figure}

We present the curves for training and validation loss and gradient norm on our pre-training data distribution in Figure \ref{fig:main_loss}. Figure \ref{fig:main_loss_train} shows that the training process of \modelname succeeds with a single, stable run without any divergence. This result is predictable with our $\mu$P hyperparameter search mentioned above. 
Figure \ref{fig:main_loss_valid} indicates that the loss curve generalizes well to validation data without saturation or overfitting. Figure \ref{fig:main_grad_norm} presents the gradient norm. 
We observe that the reduction in language modeling loss translates well into improvements on downstream tasks.

\input{tables/bpb_loss_vs_llama_v2}
\input{tables/bpb_loss_vs_llama_qwen_zh}

Language modeling is compression \cite{is_compression}. Evaluation metrics related to language perplexity (PPL) are well-known to be closely connected to compression ratio. Moreover, these metrics usually exhibit more stable scaling behavior, making them an authentic foundation of downstream task performance (which is usually measured by more complex and nonlinear metrics \cite{mirage}). For PPL-related evaluation, we use Bits-Per-Byte (BPB) \cite{paloma, pile} as our metric, which considers both per-token loss and the influence of domains and tokenizers. Specifically, on a test corpus in a certain domain, if the total loss is close, a model that tokenizes with a better compression ratio is preferred by the BPB metric.

For the English language, we break down the BPB evaluation into 6 different domains, represented by validation datasets from WebText\footnote{We use text from CommonCrawl and C4, which approximately represent the same source (broad web data).}, Github, Wikipedia, Books, ArXiv, and StackExchange, respectively. We compare with different versions of Llama, including Llama-65B, Llama2-70B, Llama3-8B, and Llama3-70B \cite{llama3}, to analyze how well \modelname fits to compress English data. Figure \ref{fig:valid_bob_loss_with_llama_4x3} illustrates the BPB trends w.r.t. to the amount of our pre-training data (in trillion tokens). As training progresses, \modelname surpasses Llama2-70B on WebText, Github, and StackExchange, outperforming Llama-65B and Llama3-8B on almost all datasets, demonstrating strong foundation abilities in English. Numerical results are presented in Table \ref{tab:bpb_loss_vs_Llama_v2}. Regarding the weighted sum of BPB, \modelname outperforms Llama-65B, Llama2-70B, Qwen1.5-72B, and Llama3-8B on both \modelname and Llama \cite{llama} weighting proportions. Note that Llama3-8B is trained on more than 15T tokens, and these results may indicate that scaling up the model size is still important, despite the rapid growth of the total amount of training data.

Similarly to English, we compute BPB across 7 domains with the corresponding Chinese validation data, namely WebText, Code, Book, World Knowledge, QA, Classical Chinese, and Professional. Results are visualized in Figure \ref{fig:valid_bob_loss_with_llama_4x3} (with ``zh'' suffix). Specific scores are provided in Table \ref{tab:bpb_loss_vs_llama_qwen_zh}. On all these validation corpora, \modelname demonstrates lower BPB than Qwen1.5-72B and the latest Llama3-70B model. Thus, we conclude that our foundation model achieves strong compression performance for Chinese without sacrificing its English language modeling abilities, and vice versa.

\section{Benchmark Evaluations}
\label{sec:benchmark-evaluation}

\subsection{English: Open LLM, HumanEval, and BBH}

\paratitle{Benchmarks.}
We evaluate \modelname on three public and widely-used English benchmarks: Open LLM Leaderboard\footnote{\url{https://huggingface.co/spaces/HuggingFaceH4/open_llm_leaderboard}.}, HumanEval \cite{humaneval}, and BIG-Bench Hard \cite{bbh}. 
\begin{itemize}[leftmargin=*]
\item Open LLM Leaderboard is hosted on Huggingface and includes 6 key tasks to measure a model's performance on a variety of areas, such as commonsense inference, knowledge capacity, truthfulness, and maths. We report our model's results with the official evaluation tools (Language Model Evaluation Harness \cite{eval-harness}). For the baseline models, we pick the results directly from the Open LLM Leaderboard. 
    \item HumanEval, introduced by OpenAI, tends to evaluate the code generation ability of language models by measuring functional correctness of docstring-prompted output. We choose the pass@5 metric as a trade-off between representing model capability and the evaluation speed.
    \item Big-Bench Hard is derived from the BIG-Bench benchmark, a diverse evaluation suite that focuses on tasks believed to be beyond the capabilities of current language models. The Big-Bench-Hard, containing 23 challenging tasks, is specifically chosen to represent areas where language models did not surpass average human-rater performance, according to prior evaluations \cite{bbhchallenging}.

\end{itemize}

\input{tables/eval_en}

\paratitle{Results.}
Table \ref{tab:eval_en} compares \modelname to the Llama series. With 52B parameters and around 1.3T English pre-training tokens, \modelname matches the overall performance of Llama-65B, which is trained on approximately 1.4T tokens. Regarding the nature of different subtasks, \modelname shows advantages over Llama-65B on GSM8K \cite{gsm8k} and HumanEval, which focus on reasoning capabilities, but performs slightly worse on some tasks that rely more heavily on knowledge. This disadvantage can potentially be mitigated with more pre-training data consumed. Besides, \modelname achieves $>90\%$ of the performances of Llama2-70B, which is larger in size and trained on a 2T token corpus.

\subsection{Chinese: OpenCompass}
\paratitle{Benchmarks.}
To measure the Chinese language and knowledge capabilities of our model, we conduct an evaluation using the OpenCompass\footnote{\url{https://opencompass.org.cn/home}.} toolkit. Specifically, we choose the following tasks to evaluate the model's performance in multiple aspects: C-Eval \cite{c-eval} and CMMLU \cite{cmmlu} (multi-subject knowledge), C3 \cite{c3} (reading comprehension), CHID \cite{chid} (Chinese culture and language understanding), and CSL \cite{csl} (keyword recognition). 

\paratitle{Results.}
Table \ref{tab:eval_cn} shows evaluation results on Chinese benchmarks. On average, \modelname achieves significantly higher scores than GPT-3.5 and comparable to GPT-4 and DeepSeek-67B \cite{deepseek}, reaching 84\% of Qwen1.5-72B's performance \cite{qwen}. Note that Qwen1.5-72B is larger in size and trained with up to 3T tokens. On CHID and CSL, \modelname shows leading performance among all the models compared. Interestingly, CHID is very specific to Chinese culture, while CSL comes from the scientific domain. This indicates \modelname's potential to both quickly adapt to a specific language and benefit from general knowledge presented in different languages.

\input{tables/eval_cn}

\subsection{Evolution of Performance during Training}
\begin{figure}
    \centering
    \includegraphics[width=1.0\textwidth]{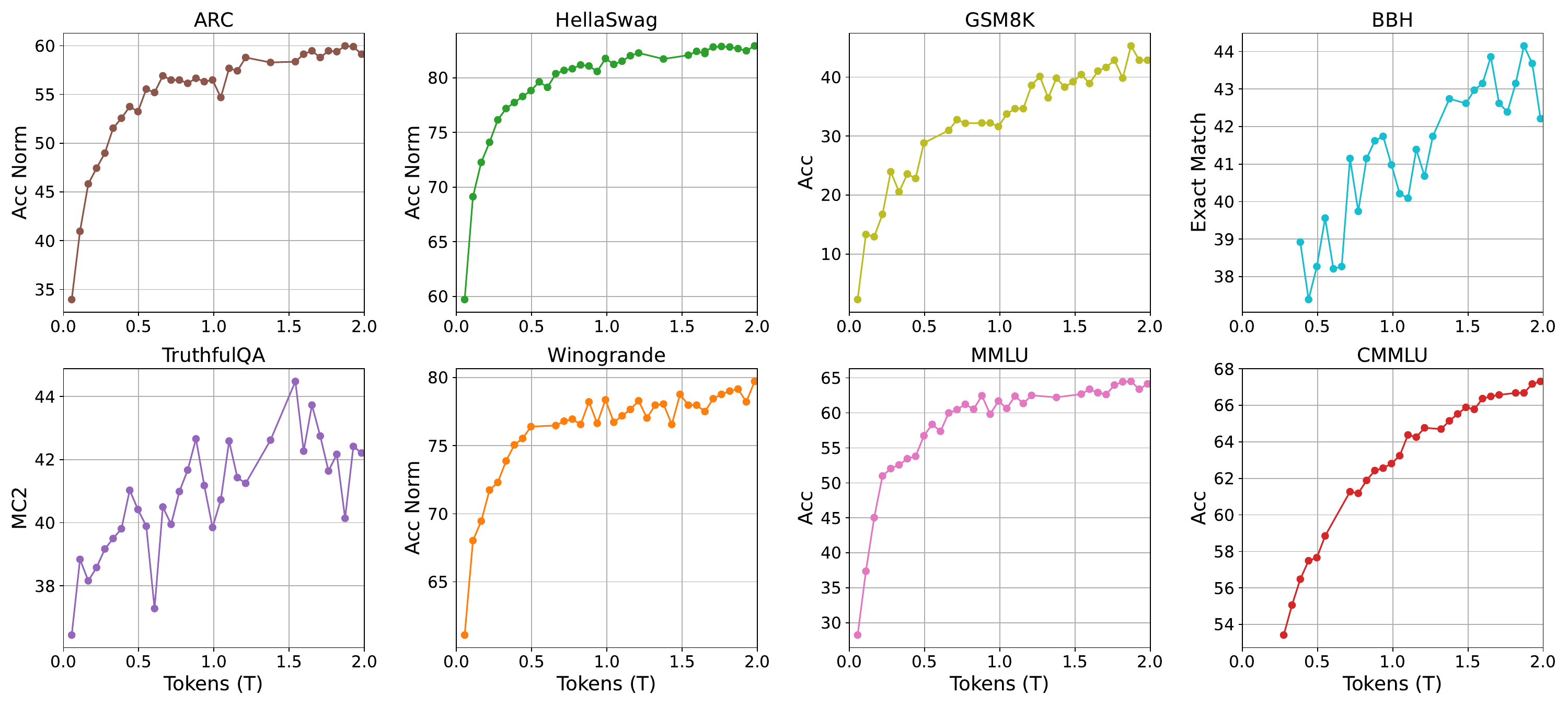}    
    \caption{\textbf{Evolution of performance evaluated by Language Model Evaluation Harness during training.} Note that we sampled 20\% examples for Hellswag and 30\% examples for MMLU considering the time cost.}
    \label{fig:evolution_performance_flm}
\end{figure}

We automatically track the evaluation scores on sampled validation data for 8 of the evaluation benchmarks, as depicted in Figure \ref{fig:evolution_performance_flm}. We observe that for all the tasks, evaluation score improves as pre-training and validation loss/BPB decreases. For knowledge-oriented English benchmarks, including ARC \cite{arc}, HellaSwag \cite{hellaswag}, Winogrande \cite{winogrande}, and MMLU \cite{mmlu}, the performances increase smoothly with more data, which is intuitive regarding the task nature. For reasoning-oriented tasks including GSM8K and BBH, we observe a sharper increase, which indicates these tasks have more complex metrics and could possibly demonstrate emergent abilities. CMMLU is a knowledge-oriented Chinese benchmark. The sharper increase in CMMLU indicates that our Chinese training data is far from saturating, and further improvement can be expected with the ongoing training process. 

\section{Lessons Learned}
\label{sec:discuss}
\paratitle{Lesson on Pre-training Data.}
We have the following observations in \modelname's pre-training process. First, as is widely known, both quality and quantity of the data are critical for pre-training; however, when there should be a trade-off between quality and quantity, data quality might be prioritized. For our project, an English-Chinese data ratio of 2:1 works better than 1:1, likely because the average quality of the Chinese Web data we have is relatively low. Second, changing the data distribution midway sometimes leads to changes in gradient norm curves and potential divergence, while maintaining a fixed distribution is more stable. Another advantage of maintaining a fixed data distribution is that it allows for safer early-stop of the $\mu$P experiments. To conclude, the data processing should be as complete as possible before the pre-training starts.

\paratitle{Lesson on Hyperparameter Search.}
We observe that $\mu$P-based methods \cite{TP5,mu-scaling} are effective and efficient in searching for the best hyperparameters and predicting the behaviors of the final large models. Specifically, prior experiences and the open-sourced learning rates are good starting points for hyperparameter search. Nevertheless, initialization standard deviation and output multipliers have more significant influences than commonly known.

\paratitle{Lesson on Loss Dynamics.}
First, the slope of the loss curve typically flattens after 500B tokens. Therefore, training should be restarted promptly if early loss values are unsatisfactory. Second, random loss spikes are common and acceptable if the gradient norm curve looks normal. We observe that our model recovers from all the spikes in the pre-training process, unlike the early open-sourced endeavors \cite{OPT, palm2, glm-130b}. We speculate that modern Llama-like structures, especially those with non-bias designs and truncated normal initialization, combined with effective hyperparameter search, provide decent robustness against loss spikes. Another type of spike corresponds to consistent loss increases, which can be identified early with $\mu$P and avoided before the training begins.

\paratitle{Lesson on Gradient Norm.}
The early gradient norm curves are not strong indicators of training stability. In hyperparameter search, we observe divergence following various gradient curve patterns, yet with higher divergence probabilities associated with continuously increasing gradient trends.

\section{Related Work}
\label{sec:related}
The idea of large foundation models originates from unsupervised pre-training with Transformer-based \cite{vaswani2017attention} architectures. Well-known examples of early foundation models include Bert \cite{bert}, GPT-2 \cite{radford2019language}, and T5 \cite{t5}. GPT-3 \cite{gpt3} increases the model size to 175B and observes decent few-shot and zero-shot reasoning capabilities, which encourages a series of efforts to scale up foundation models \cite{OPT,bloom, palm2, glm-130b}. Research on scaling laws \cite{scaling-law, scaling-law-2, chinchilla, mu-scaling} sheds light on the predictable trends of model performance when the parameter number increases. On the other hand, other works explore the emergent abilities \cite{emergent, u-shape, mirage} and their relationships to evaluation metrics and task nature.

The Llama series \cite{llama, llama-2, llama3} is well-known for its contributions to open-sourced large language models, and is widely regarded as a strong baseline for foundation model evaluation. Falcon \cite{falcon} explores data processing of publicly available pre-training corpora. Mistral \cite{mistral} and Gemma \cite{gemma} release 7B-scaled models that are trained with more data and incorporated with advanced designs. For the Chinese community, Qwen \cite{qwen}, Baichuan \cite{baichuan}, Yi \cite{yi}, and DeepSeek \cite{deepseek} represent efforts in multilingual foundation model pre-training and open-sourcing. FLM-101B \cite{flm101b} studies methodologies for training large foundation models under limited budgets.

InstructGPT \cite{instructgpt} establishes the paradigm of aligning large foundation models with human preferences. Widely used approaches include supervised fine-tuning (SFT) \cite{alpaca, wizardlm} and Reinforcement Learning from Human Feedback (RLHF) \cite{PPO}, among others \cite{DPO}. Aligning techniques turn foundation models into dialogue agents, which form the core of AI assistants in commercial use. Closed-source dialogue agents are represented by GPT-4 \cite{GPT-4}, Claude \cite{claude}, Grok \cite{grok}, and Gemini \cite{gemini}. Open-sourced chat models include Zephyr \cite{zephyr} and ChatGLM \cite{chatglm}, among the large number of human-aligned versions of the open foundation models mentioned above.

\section{Conclusions and Future Work}
\label{sec:con}
In this report, we introduce \modelname, an open multilingual foundation model. With 52B parameters and 2T training tokens, \modelname matches the performance of larger models trained with more data, in both multilingual language modeling capabilities and benchmark evaluations. The pre-training procedure of \modelname features a high success rate and low carbon footprint. We open-source the model weights as well as technical details and training dynamics. We hope this work will catalyze the growth of open-sourced LLM communities and reduce the trial-and-error cycles to train LLMs with more than 50B parameters. Note that although efforts are made to filter out harmful contents in the training data, such kind of outputs could still potentially be elicited from the released model, which does not represent the opinions of the authors or entities involved. 

For future work, we plan to continue enhancing the capabilities of \modelname to facilitate broader application, as well as to develop efficient training techniques to explore the unmanned deep space of larger-scaled dense models. 

\section*{Acknowledgments}

This work is supported by the National Science and Technology Major Project (No. 2022ZD0116300) and the National Science Foundation of China (No. 62106249).
We would like to thank Boya Wu, Li Du, Quanyue Ma, Hanyu Zhao, Shiyu Wu and Kaipeng Jia for their help on data, Hailong Qian, Jinglong Li, Taojia Liu, Junjie Wang, Yuanlin Cai, Jiahao Guo, Quan Zhao, Xuwei Yang, Hanxiao Qu, Yan Tian, and Kailong Xie for their help on computational resources, and all other colleagues' strong support for this project.

\bibliographystyle{plain}
\bibliography{custom}

\end{document}

%% file: tables/train_data_stat.tex
\begin{table}[htbp]
\centering
\caption{\textbf{Pre-training data.} For each subset of our 2T pre-training tokens, we detail the language, the sampling proportion, the number of epochs completed during training, and the disk size.}
\scalebox{0.77}{
\begin{tabular}{@{}lcccr@{}}
\toprule
\textbf{Domain} & \multicolumn{1}{c}{\textbf{Language}} & \multicolumn{1}{c}{\textbf{Sampling Prop.}} & \multicolumn{1}{c}{\textbf{Epochs}} & \multicolumn{1}{r}{\textbf{Disk Size}} \\ \midrule
WebText           & en, zh        & \enspace 75.21\%        & 1.0    & 5.9 TB    \\
Code              & code, zh      & \enspace 9.81\%         & 1.0    & 528.1 GB  \\
Book              & en, zh        & \enspace 7.17\%         & 0.8    & 647.6 GB  \\
WorldKnowledge   & multi., en, zh & \enspace 2.87\%         & 2.5    & 67.5 GB   \\
QA                & en, zh        & \enspace 2.12\%         & 1.0    & 159.2 GB  \\
AcademicPaper    & en            & \enspace 0.99\%         & 1.0    & 54.4 GB   \\
Profession-Law               & zh            & \enspace 1.04\%         & 1.0    & 84.2 GB   \\
Profession-Math              & math          & \enspace 0.62\%         & 2.0    & 6.1 GB    \\
Profession-Patent            & zh            & \enspace 0.14\%         & 1.0    & 10.4 GB   \\
Profession-Medical           & zh            & \enspace 0.02\%         & 1.0    & 1.2 GB    \\ 
ClassicalChinese & zh            & \enspace 0.02\%         & 2.5    & 0.5 GB    \\
\bottomrule
\end{tabular}
}
\label{train_data_stat}
\end{table}

%% file: tables/model_architecture_parameters.tex
\begin{table}
  \centering
  \caption{\textbf{Detailed model architecture.} The model configuration of $\text{\modelname}_{\mu\text{P}}$ is a reduced version of \modelname with a smaller hidden size.}
  \scalebox{0.7}
  {
    \begin{tabular}{l|ccccccc}
    \toprule
    Models & \multicolumn{1}{l}{Layer Num} & \multicolumn{1}{l}{Attention Heads} & \multicolumn{1}{l}{Hidden Size} & \multicolumn{1}{l}{FFN Hidden Size} & \multicolumn{1}{l}{Vocab Size} & \multicolumn{1}{l}{Context Length} & \multicolumn{1}{l}{Params Size~(M)} \\
    \midrule
    \modelname & 64    & 64    & 8,192  & 21,824 & 80,000 & 4,096  & 52,850 \\
    $\text{\modelname}_{\mu\text{P}}$ & 64    & 4     & 512   & 1,344  & 80,000 & 4,096  & 283 \\
    \bottomrule
    \end{tabular}%
    }
  \label{tab:model_architecture_parameters}%
\end{table}%

%% file: tables/tokenizer_compression_rate.tex
\begin{table}
  \centering
  \caption{\textbf{Tokenizer compression ratio.} Tokenizer Compression Ratio is defined as the ratio of token length to the original UTF-8 text length. Smaller values indicate better compression. We report the compression ratios of GPT-4, Llama1/2, Llama3, and \modelname on various domains in our training set, as well as the weighted average.}
  \scalebox{0.77}
  {
    \begin{tabular}{lcccccccc}
    \toprule
    \multicolumn{1}{c}{\multirow{2}[4]{*}{Tokenizer}} & \multirow{2}[4]{*}{Vocab Size} & \multicolumn{7}{c}{Compression Rate} \\
\cmidrule{3-9}          &       & \multicolumn{1}{l}{English} & \multicolumn{1}{l}{Chinese} & \multicolumn{1}{l}{Classical Chinese} & \multicolumn{1}{l}{Code} & \multicolumn{1}{l}{Multilingual} & \multicolumn{1}{l}{Mathematical} & \multicolumn{1}{l}{Weighted Avg.} \\
    \midrule
    GPT-4 & 100k  & 0.221 & 0.420 & 0.478 & 0.267 & 0.303 & 0.508 & 0.291 \\
    Llama1/2 & 32k   & 0.262 & 0.515 & 0.558 & 0.367 & 0.314 & 0.974 & 0.356 \\
    Llama3 & 128k  & 0.220 & 0.294 & 0.353 & 0.267 & 0.274 & 0.508 & 0.251 \\
    \modelname & 80k   & 0.248 & 0.235 & 0.307 & 0.363 & 0.340 & 0.965 & 0.261 \\
    \bottomrule
    \end{tabular}%
    }
  \label{tab:tokenizer_compression_rate}%
\end{table}%

%% file: tables/hyperparameters.tex
\begin{table}
  \centering
  \caption{\modelname Training Hyperparameters.}
    \begin{tabular}{lr|lr}
    \toprule
    \multicolumn{2}{c|}{Searched Hyperparameters} & \multicolumn{2}{c}{Non-Searched Hyperparameters} \\
    \midrule
    Learning Rate & 1.5e-4 & LR Schedule Type & cosine \\
    Matrix Learning Rate & 1.5e-4 & LR Schedule (tokens) & 2.5T \\
     Minimum Learning Rate & 1.5e-5  & Warmup Step & 2,000 \\
     Standard Deviation & 4e-3 & Clip Grad & 1.0 \\
     Matrix Standard Deviation & 4.242e-3& Weight Decay & 0.0 \\
   Input Mult & 1.0   & Batch Size (tokens) & 5,505,024 \\
    Output Mult & 3.125e-2   & RoPE Theta & 10,000 \\
    \bottomrule
    \end{tabular}%
  \label{tab:hyperparameters}%
\end{table}%

%% file: tables/bpb_loss_vs_llama_v2.tex
\begin{table}[t]
  \centering
  \caption{\textbf{BPB of \modelname, Llama family models, and Qwen1.5-72B on English datasets.} BPB is computed for 6 dataset categories, with weighted sum results based on Llama \cite{llama} and \modelname training data configurations. The best results are in boldface and second-best underlined.}
  \scalebox{0.75}{
  \begin{threeparttable}
    \begin{tabular}{cl|cccccc|cc}
    \toprule
    \multicolumn{2}{c|}{\multirow{2}[2]{*}{Model}} & \multicolumn{1}{r}{\multirow{2}[2]{*}{WebText}}  & \multicolumn{1}{r}{\multirow{2}[2]{*}{Github}} & \multicolumn{1}{r}{\multirow{2}[2]{*}{Wikipedia}} & \multicolumn{1}{r}{\multirow{2}[2]{*}{Book}} & \multicolumn{1}{r}{\multirow{2}[2]{*}{ArXiv}} & \multicolumn{1}{r|}{\multirow{2}[2]{*}{StackExchange}} & \multicolumn{2}{c}{Weighted Sum} \\
    \multicolumn{2}{c|}{} &  &       &       &       &       &       & L-Prop.\tnote{1} & F-Prop.\tnote{2} \\
    \midrule
    \multirow[c]{5}{*}{Loss} & Llama-65B & 1.650 & 0.543 & 1.297 & 1.791 & 1.205 & 1.293 & 1.572 & 1.485 \\
          & Llama2-70B & 1.588 & 0.471 & 1.198 & 1.695 & 1.103 & 1.220 & 1.506 & 1.418 \\
          & Llama3-70B & 1.729 & 0.597 & 1.300 & 1.886 & 1.042 & 1.388 & 1.642 & 1.556 \\
          & Qwen1.5-72B & 1.996 & 0.592 & 1.433 & 2.107 & 1.111 & 1.393 & 1.878 & 1.773 \\
          & \modelname (52B) & 1.598 & 0.314 & 1.163 & 1.843 & 1.153 & 1.193 & 1.512 & 1.411 \\
    \midrule
    \multirow[c]{5}{*}{BPB} & Llama-65B & 0.615 & 0.286 & 0.595 & 0.710 & 0.590 & 0.570 & 0.602 & 0.574 \\
          & Llama2-70B & 0.592 & 0.249 & \underline{0.544} & \underline{0.672} & 0.540 & 0.538 & 0.576 & 0.547 \\
          & Llama3-70B & \textbf{0.542} & \underline{0.229} & \textbf{0.513} & \textbf{0.633} & \textbf{0.479} & \textbf{0.497} & \textbf{0.528} & \textbf{0.502} \\
          & Qwen1.5-72B & 0.642 & 0.234 & 0.601 & 0.717 & \underline{0.521} & \underline{0.515} & 0.620 & 0.586 \\
          & \modelname (52B) & \underline{0.562} & \textbf{0.164} & 0.570 & 0.700 & 0.567 & 0.531 & \underline{0.550} & \underline{0.516} \\
    \bottomrule
    \end{tabular}%
            \begin{tablenotes}
            \item[1] L-Prop.~(Llama \cite{llama} Proportion): 82\% : 4.5\% : 4.5\% : 4.5\% : 2.5\% : 2.0\%.
            \item[2] F-Prop.~(\modelname Proportion): 75.17\% : 13.48\% : 3.56\% : 5.26\% : 1.46\% : 1.07\%. 
        \end{tablenotes}
    \end{threeparttable}
}
  \label{tab:bpb_loss_vs_Llama_v2}%
\end{table}%

%% file: tables/bpb_loss_vs_llama_qwen_zh.tex
\begin{table}[t]
  \centering
  \caption{\textbf{BPB of \modelname, Llama family models and Qwen1.5-72B, on Chinese datasets.} BPB is computed for 7 dataset categories, with direct average and weighted sum results based on \modelname training data distributions.}
  \scalebox{0.74}
  {
  \begin{threeparttable}
    \begin{tabular}{cl|ccccccc|cc}
    \toprule
    \multicolumn{2}{c|}{\multirow{2}[2]{*}{Models}} & \multirow{2}[2]{*}{WebText} & \multirow{2}[2]{*}{Code} & \multirow{2}[2]{*}{Book} & World & \multirow{2}[2]{*}{QA} & Classical & \multirow{2}[2]{*}{Professional} & Direct & Weighted\tnote{1} \\
    \multicolumn{2}{c|}{} &       &       &       &  Knowledge &       & Chinese &       & Average & Sum \\
    \midrule
    \multirow[c]{5}{*}{Loss} & Llama-65B & 1.773 & 1.236 & 2.029 & 1.586 & 2.076 & 2.819 & 1.215 & 1.819 & 1.782 \\
          & Llama2-70B & 1.419 & 1.019 & 1.542 & 1.189 & 1.681 & 2.233 & 0.896 & 1.426 & 1.414 \\
          & Llama3-70B & 2.152 & 1.264 & 2.210 & 1.722 & 2.568 & 2.844 & 1.109 & 1.981 & 2.114 \\
          & Qwen1.5-72B & 2.260 & 1.405 & 2.520 & 1.751 & 2.888 & 2.748 & 0.908 & 2.069 & 2.243 \\
          & \modelname (52B) & 1.923 & 1.096 & 2.135 & 1.612 & 2.530 & 2.144 & 0.846 & 1.755 & 1.913 \\
    \midrule
    \multirow[c]{5}{*}{BPB} & Llama-65B & 1.325 & 0.744 & 1.503 & 1.161 & 1.528 & 2.280  & 0.919 & 1.351 & 1.326 \\
          & Llama2-70B & 1.060 & 0.614 & 1.142 & 0.869 & 1.237 & 1.811 & 0.678 & 1.059 & 1.052 \\
          & Llama3-70B & 0.913 & \underline{0.498} & 0.943 & 0.752 & 1.063 & 1.458 & 0.485 & 0.873 & 0.897 \\
          & Qwen1.5-72B & \underline{0.759} & 0.537 & \underline{0.871} & \underline{0.663} & \underline{0.951} & \underline{1.237} & \underline{0.329} & \underline{0.764} & \underline{0.759} \\
          & \modelname (52B) & \textbf{0.643} & \textbf{0.478} & \textbf{0.741} & \textbf{0.619} & \textbf{0.831} & \textbf{0.949} & \textbf{0.290} & \textbf{0.650} & \textbf{0.646} \\
    \bottomrule
    \end{tabular}%
    \begin{tablenotes}
            \item[1] \modelname training set Proportion: 76.60\% : 1.91\% : 11.61\% : 1.44\% : 4.50\% : 0.07\% : 3.87\%. 
    \end{tablenotes}
    \end{threeparttable}
    }
  \label{tab:bpb_loss_vs_llama_qwen_zh}%
\end{table}%

%% file: tables/eval_en.tex
\begin{table}[h]
\centering
\caption{Performance of \modelname and baselines on English benchmarks.}
\footnotesize
\scalebox{0.77}
{

\begin{tabular}{@{}lccccccccc@{}}
\toprule
\multicolumn{1}{l|}{\multirow{2}{*}{Model}} & \multirow{2}{*}{Average} & ARC                  & HellaSwag            & MMLU                 & TruthfulQA           & WinoGrande            & GSM8K                & HumanEval            & BBH                  \\
\multicolumn{1}{l|}{}                       &                          & 25-shot              & 10-shot              & 5-shot               & zero-shot            & 5-shot               & 5-shot               & zero-shot            & 3-shot               \\ \midrule
\multicolumn{1}{l|}{Llama2-70B}             & 63.39                    & 67.32                & 87.33                & 69.83                & 44.92                & 83.74                & 54.06                & 46.95                & 52.94                \\
\multicolumn{1}{l|}{Llama2-13B}             & 50.29                    & 59.39                & 82.13                & 55.77                & 37.38                & 76.64                & 22.82                & 28.66                & 39.52                \\
\multicolumn{1}{l|}{Llama-65B}              & 56.98                    & 63.48                & 86.09                & 63.93                & 43.43                & 82.56                & 37.23                & 33.54                & 45.54                \\
\multicolumn{1}{l|}{Llama-13B}              & 46.20                    & 56.23                & 80.93                & 47.67                & 39.48                & 76.24                & \enspace 7.58                 & 23.78                & 37.72                \\
\multicolumn{1}{l|}{Tele-FLM~(52B)}              & 56.60     & 59.47   & 82.25     & 64.00     & 43.09      & 79.40      & 45.19  & 34.76     & 44.60   \\
\bottomrule

\end{tabular}
}
\label{tab:eval_en}
\end{table}

%% file: tables/eval_cn.tex
\begin{table}[t]
\centering
\caption{\textbf{Performance of \modelname and baselines on Chinese benchmarks.} The results of Qwen1.5-72B and our \modelname are locally computed with the OpenCompass toolkit, while other results are picked from OpenCompass leaderboard. }
\footnotesize
\scalebox{0.9}
{

\begin{tabular}{@{}lcccccc@{}}
\toprule
Model        & Average & C-Eval & CMMLU & C3    & CHID  & CSL   \\ \midrule
GPT-4        & 76.64   & 69.90  & 71.00 & 95.10 & 82.20 & 65.00 \\
GPT-3.5      & 61.86   & 52.50  & 53.90 & 85.60 & 60.40 & 56.90 \\
Qwen1.5-72B  & 80.45   & 83.72  & 83.09 & 81.86 & 91.09 & 62.50 \\
Qwen-72B     & 83.00   & 83.30  & 83.60 & 95.80 & 91.10 & 61.20 \\
DeepSeek-67B & 73.46   & 66.90  & 70.40 & 77.80 & 89.10 & 63.10 \\
Tele-FLM~(52B)     & 71.13   & 65.48  & 66.98 & 66.25 & 92.57 & 64.38 \\ \bottomrule
\end{tabular}
}
\label{tab:eval_cn}
\end{table}